\documentclass[runningheads]{llncs}
\usepackage[T1]{fontenc}
\usepackage{multirow}
\usepackage{graphicx,verbatim}
\usepackage{amsmath}
\usepackage{amssymb}
\usepackage{hyperref}
\usepackage{algorithm}
\usepackage{algpseudocode}
\usepackage{siunitx}
\usepackage{tikz}
\usepackage{booktabs}
\usepackage{placeins}
\newcommand{\bftab}{\fontseries{b}\selectfont}

\usepackage{color}

\begin{document}
\title{3D CBCT Artefact Removal Using Perpendicular Score-Based Diffusion Models}
\titlerunning{Artefact Removal Using Score-based Diffusion Models}

\author{Susanne Schaub\inst{1}\orcidID{0009-0008-6278-4855
} \and Florentin Bieder \inst{1}\orcidID{0000-0001-9558-0623
} \and Matheus L. Oliveira \inst{2}\orcidID{0000-0002-8054-8759
} \and Yulan Wang\inst{3,5}\orcidID{0000-0003-0502-6931}
\and Dorothea Dagassan-Berndt\inst{4}\orcidID{0000-0003-1109-8763}
\and Michael M. Bornstein\inst{5}\orcidID{0000-0002-7773-8957}
\and Philippe C. Cattin\inst{1} \orcidID{0000-0001-8785-2713} }

\authorrunning{S. Schaub et al.}

\institute{Department of Biomedical Engineering, University of Basel, Allschwil, Switzerland
\email{s.schaub@unibas.ch}
\and
Piracicaba Dental School, University of Campinas, Piracicaba, Brazil
\and
State Key Laboratory of Oral \& Maxillofacial Reconstruction and Regeneration, Key Laboratory of Oral Biomedicine Ministry of Education, Hubei Key Laboratory of Stomatology, School \& Hospital of Stomatology, Wuhan University, Wuhan, China
\and
Dental Imaging, University Center for Dental Medicine, University of Basel, Basel, Switzerland
\and
Department of Oral Health \& Medicine, University Center for Dental Medicine, University of Basel, Basel, Switzerland
}
\maketitle              
\begin{abstract}
Cone-beam computed tomography (CBCT) is a widely used 3D imaging technique in dentistry, offering high-resolution images while minimising radiation exposure for patients. However, CBCT is highly susceptible to artefacts arising from high-density objects such as dental implants, which can compromise image quality and diagnostic accuracy. To reduce artefacts, implant inpainting in the sequence of projections plays a crucial role in many artefact reduction approaches. Recently, diffusion models have achieved state-of-the-art results in image generation and have widely been applied to image inpainting tasks. However, to our knowledge, existing diffusion-based methods for implant inpainting operate on independent 2D projections. This approach neglects the correlations among individual projections, resulting in inconsistencies in the reconstructed images. To address this, we propose a 3D dental implant inpainting approach based on perpendicular score-based diffusion models, each trained in two different planes and operating in the projection domain. The 3D distribution of the projection series is modelled by combining the two 2D score-based diffusion models in the sampling scheme. Our results demonstrate the method's effectiveness in producing high-quality, artefact-reduced 3D CBCT images, making it a promising solution for improving clinical imaging. Our code is publicly available at \url{https://github.com/SusanneSchaub/TPDM_Implant_Inpainting}.
\end{abstract}
\section{Introduction}
High-density objects, such as dental implants and endodontic filling materials, introduce artefacts in cone-beam computed tomography (CBCT) images. These artefacts manifest as distortions in the reconstructed images that do not correspond to the actual scanned object, leading to reduced image quality and potentially compromising image analysis and diagnostic accuracy.
This work focuses on metal artefact reduction in CBCT imaging, a widely used 3D radiographic technique in fields such as implant dentistry and image-guided radiation therapy. This modality can produce high-quality images while reducing radiation exposure compared to fan-beam CT. 
The field of generative modelling has witnessed remarkable advancements in recent years, driven by the development of powerful deep learning frameworks. Among these, diffusion models have emerged as a promising model family for state-of-the-art data generation~\cite{friedrich2024wdm,wang20243d}, outperforming models such as generative adversarial networks and variational autoencoders in various tasks. Several deep learning-based methods have been developed for artefact removal, operating in the reconstructed image domain~\cite{oliveira2025development,zhang2018convolutional}, the projection domain~\cite{gottschalk2023dl,mei2023metal}, or in the dual domain~\cite{lin2019dudonet,tang2023improved}. Inpainting approaches in the projection domain offer a key advantage by preventing artefacts from forming in the first place, unlike image-domain methods, where artefacts have already degraded the data, potentially leading to irreversible information loss. 
\begin{figure}[h!]
\centering
\includegraphics[width=\textwidth, height=5.0cm, keepaspectratio]{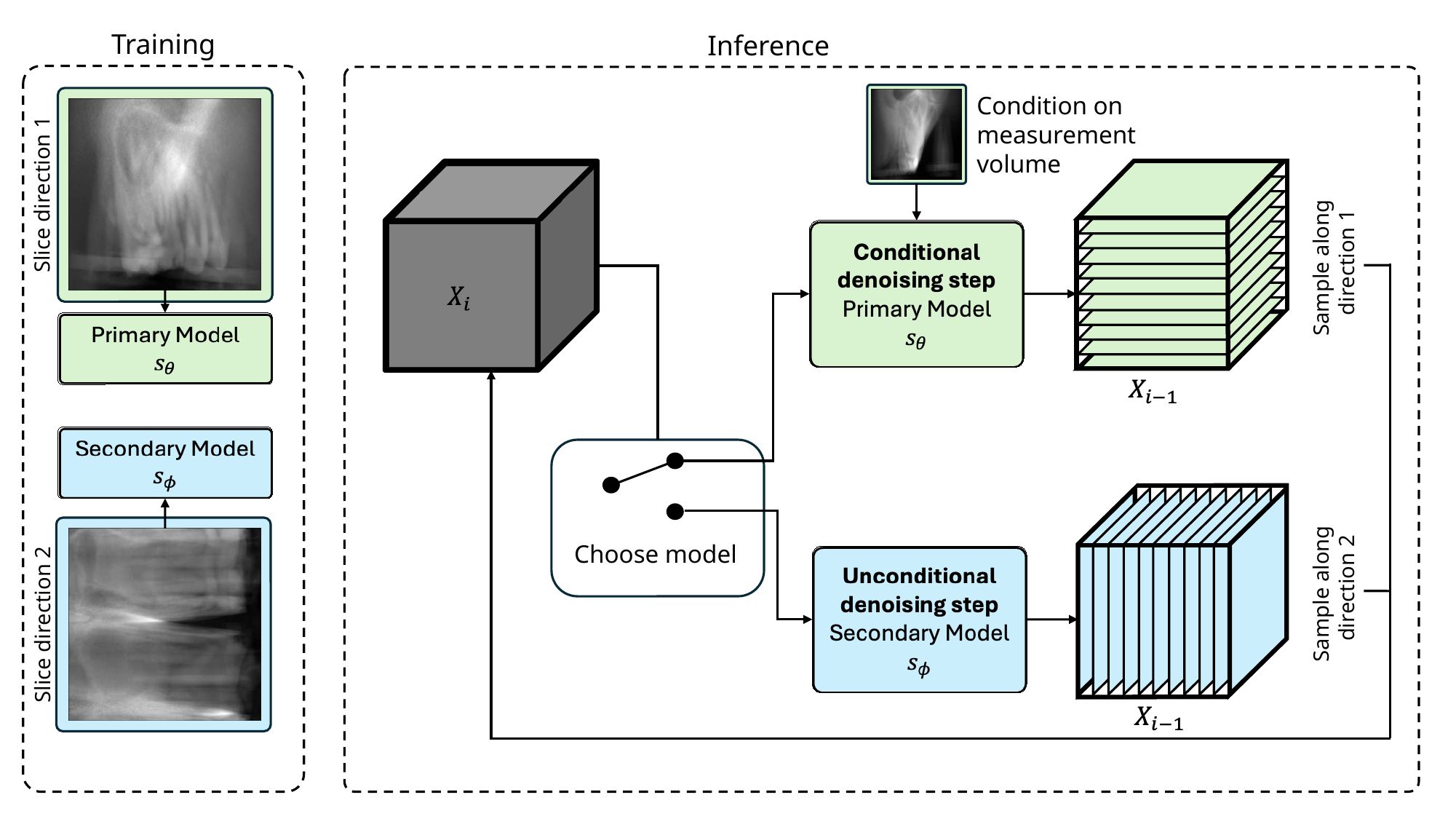}
\caption{A primary diffusion model is trained on projections along the original projection direction, while a secondary model is trained on perpendicular projections. During sampling, the algorithm alternates between the two models to generate a coherent 3D image.} \label{fig1}
\end{figure}
\FloatBarrier
To our knowledge, diffusion model-based approaches for implant inpainting in the projection domain have only been proposed for 2D data~\cite{karageorgos2024denoising,wang2024metal}. These models suffer from inconsistencies in the volumetric reconstructions. To address this issue, we propose a 3D implant inpainting approach utilising two perpendicular score-based diffusion models, each trained on projection series split along different orientations. These models are combined in the sampling scheme to acquire the 3D projection stacks. Training full 3D models is typically computationally expensive and requires large amounts of data. By training our models on 2D projections instead, we effectively circumvent these challenges while maintaining a high reconstruction quality in 3D.

A recent study showed that CBCT scanner-integrated metal artefact reduction algorithms are not effective in reducing artefacts induced by implants located in the exomass~\cite{candemil2019metal}. The exomass is the zone between the source and the detector but outside of the field of view (FOV). The use of small-sized FOVs is recommended whenever possible, as it enhances image sharpness while minimising the radiation dose to patients. However, with smaller FOVs, dental implants may fall within the exomass. We trained our models on both large and small FOVs, enabling them to handle both settings.
\section{Background}
\subsubsection{Score-Based Diffusion Models}
Score-based generative models introduced by Song et al.~\cite{song2020score} effectively leverage score functions to model complex data distributions and have gained attention for their ability to solve various inverse problems without training the model for a specified inverse problem task. 
\subsubsection{Diffusion Posterior Sampling} The diffusion posterior sampling (DPS) technique introduced by Chung et al.~\cite{chung2022diffusion} is a method to solve noisy (non-)linear inverse problems by approximating the posterior sampling. Given an inverse problem
\begin{equation}
\mathbf{y} = \mathcal{A}\left( \mathbf{x_0} \right) + \mathbf{n},
\end{equation} where $\mathcal{A}: \mathbb R^{d_1 \times d_2} \to \mathbb R^{d_1' \times d_2'}$
is the forward measurement function and $\mathbf{n}$ is the measurement noise, Bayes' rule can be applied to obtain
\begin{align}
\begin{split}
\nabla_{\mathbf{x_{t}}} \log p \left( \mathbf{x_t} | \mathbf{y} \right) &=\nabla_{\mathbf{x_{t}}} \log p \left( \mathbf{x_t} \right) + \nabla_{\mathbf{x_{t}}} \log p \left( \mathbf{y} | \mathbf{x_t} \right) 
\\ &\simeq \mathbf{s_{\theta}} \left( \mathbf{x_t} , t \right) + \nabla_{\mathbf{x_{t}}} \log p \left( \mathbf{y} | \mathbf{x_t} \right).
\end{split}
\end{align} 
The approximation of the first term is given by the score-based model $\mathbf{s_{\theta}}$. The second term is difficult to acquire in closed form since there is no explicit relationship between $\mathbf{x_t}$ and $\mathbf{y}$. The approximation
\begin{equation}
\nabla_{\mathbf{x_{t}}} \log p \left( \mathbf{y} | \mathbf{x_t} \right) \simeq \nabla_{\mathbf{x_{t}}} \log p \left( \mathbf{y}| \mathbf{\hat{x}_0} \left( \mathbf{x_t} \right) \right),
\end{equation} where
\begin{equation}
\mathbf{\hat{x}_0} \left( \mathbf{x_t} \right) = \mathbb E \left[ \mathbf{x_0} | \mathbf{x_t} \right] = \mathbf{x_t} + \sigma^{2} \left( t \right) \nabla_{\mathbf{x_{t}}} \log p \left( \mathbf{x_t} \right)
\end{equation} has been proposed in~\cite{chung2022diffusion}. Therefore, the posterior can be approximated with 
\begin{equation}
\nabla_\mathbf{{x_{t}}} \log p \left( \mathbf{x_t} | \mathbf{y} \right) \simeq \mathbf{s_{\theta}} \left( \mathbf{x_t} , t \right) - \lambda \nabla_{\mathbf{x_{t}}} \left\Vert \mathcal{A}\left( \mathbf{\hat{x}_0} \left( \mathbf{x_t} \right) \right) - \mathbf{y} \right\Vert^{2}_{2} .
\end{equation}
The hyperparameter $\lambda$ determines the extent of conditioning on $\mathbf{y}$.
\section{Method}
\begin{figure}
\centering
\includegraphics[width=\textwidth,height=5cm, keepaspectratio]{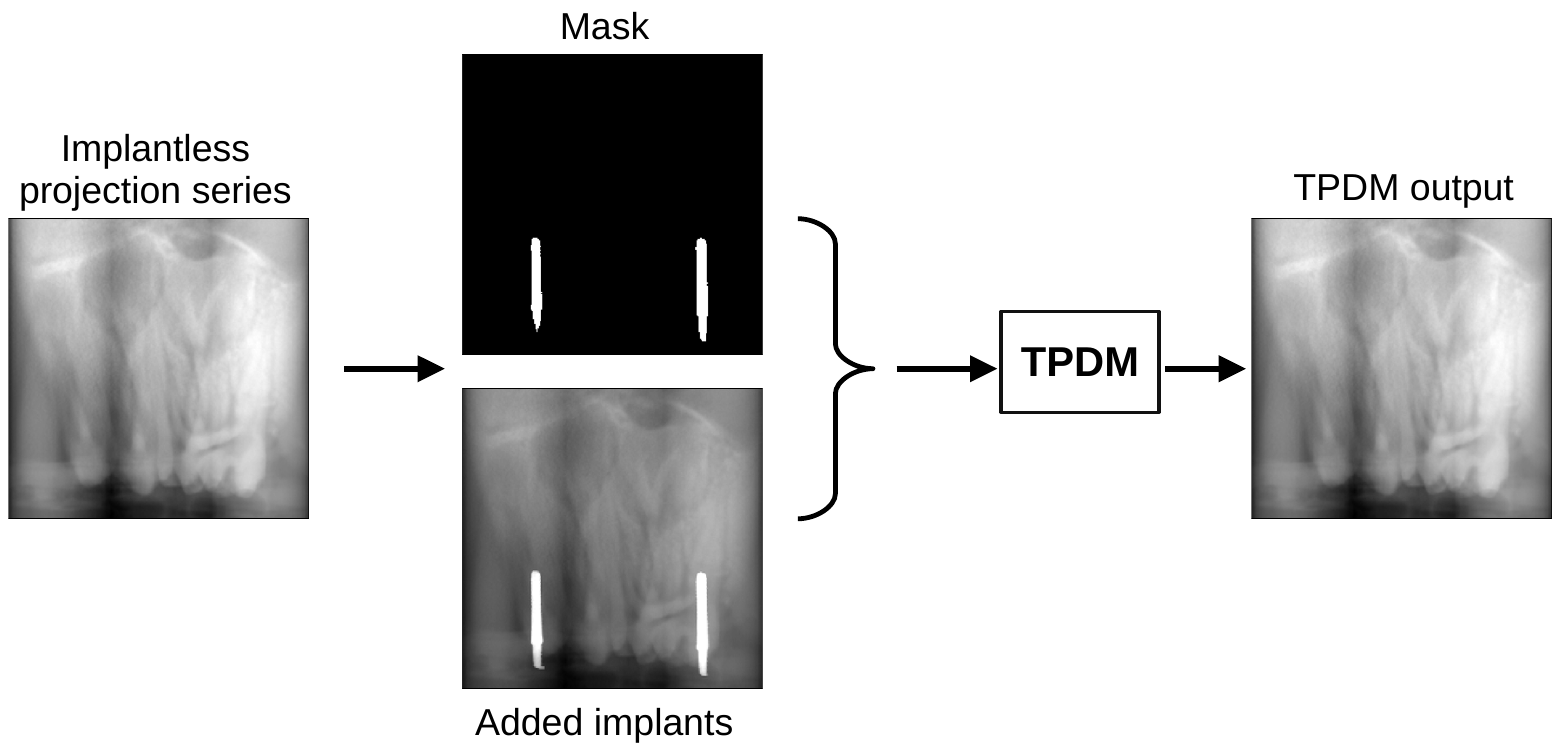}
\caption{We computed the projection series from our implant-free CBCT reconstructions. Implants were added with a mask, which serves as input to the TPDM.} \label{figwork}
\end{figure}

Lee et al.~\cite{lee2023improving} proposed modelling the 3D data distribution $p_{\theta, \phi} $ as a product of two distributions, given by
\begin{equation}
p_{\theta, \phi} \left( \mathbf{x} \right) = q_{\theta}^{\alpha} \left( \mathbf{x} \right) q_{\phi}^{\beta} \left( \mathbf x \right) / Z,
\end{equation}
where $q_{\theta} $ is the distribution modelled by the primary model, $q_{\phi}$ the distribution modelled by the secondary model, $\alpha$ and $\beta$ are constants, and $Z$ is a normalisation function.
The distributions $q_\theta$ and $q_\phi$ are parametrized by $\theta$ and $\phi$, respectively. Both distributions $q_\theta$ and $q_\phi$ are a decomposition of 2D contributions
\begin{equation}
\begin{split}
& \nabla_{\mathbf{x_{t}}} \log p_{\theta, \phi} \left( \mathbf{x_{t}} \right) \simeq  \alpha \sum_{i=1}^{d_3} \mathbf{s_{\theta}}^{3D} \left( \mathbf{x_{t}} \left[ :,:,i \right] \right) + \beta \sum_{i=1}^{d_1} \mathbf{s_{\phi}}^{3D} \left( \mathbf{x_{t}} \left[ i,:,: \right] \right), \\
& \mathbf{s_{\theta}}^{3D} \left( \mathbf{x_{t}} \left[ :,:,i \right] \right)_{\left[ :,:,j \right]}  =  \delta_{ij} \mathbf{s_{\theta}} \left( \mathbf{x_{t}} \left[ :,:,i \right] \right) ,
\end{split}
\end{equation}
where $\mathbf{s_{\theta}}$ denotes the 2D primary score-based diffusion model and $\mathbf{s_{\phi}}$ the secondary 2D score-based diffusion model. To indicate partial indexing along a specific axis, we borrow the square bracket notation from Python. A schematic overview of the method is shown in Fig.~\ref{fig1}. The primary diffusion model was trained on 2D projections along the original projection direction, while the secondary model was trained on images perpendicular to this direction. During sampling with the two perpendicular diffusion models (TPDM), we iteratively progressed through all diffusion steps and determined at each step whether to use the primary or secondary model. Depending on the chosen model, we sampled the volume along either the primary or secondary model's training direction. Conditioning on the measurement volume occurred only when using the primary model. The sampling scheme is shown in Algorithm~\ref{alg:cap}. In our case, the measurement volume $\mathbf{Y}$ is the stacked projection series with added implants, and the measurement function $\mathcal{A(\cdot)}$ is defined by applying the implant mask to it. The input to the TPDM models is shown in Fig.~\ref{figwork}.
\begin{algorithm}[h!]
\small
\caption{Sampling scheme}\label{alg:cap}
\begin{algorithmic}
\Require $ \mathbf{Y}\in\mathbb{R}^{d_1 \times d_2 \times d_3}$, $\mathcal{A(\cdot)}$, $\mathbf{s_{\theta}}$, $\mathbf{s_{\phi}}$, $T$, $K$, $\lambda$, $\{\sigma_i\}_{0}^{1}$, $X_T \sim \mathcal{N}(\mathbf{0},\sigma_i^{2}\mathbf{I}) \in \mathbb{R}^{d_1 \times d_2 \times d_3}$
\For{$i$ in $T-1:0 $}

$t \gets \frac{i}{T}$
\If{$\mod (i,K) \neq 0$} 
    \For{$j$ in $1:d_3$} \Comment{Primary model chosen}
        \State $\mathbf{x} \gets \mathbf{X_{i+1}}[:,:,j]$
        \State $\mathbf{y} \gets \mathbf{Y}[:,:,j]$
        \State $\mathbf{\hat{x}_0} \gets \mathbf{x} + \sigma_{t}^2  \cdot$ $\mathbf{s_{\theta}}(\mathbf{x},t)$
        \State $\mathbf{x'} \gets \texttt{PC sampling} (\mathbf{x}, \mathbf{s_{\theta}}, \sigma_t, t)$
        \Comment{Predictor-Corrector sampling step~\cite{song2020score}}
        \State $\mathbf{x''} \gets \mathbf{x'} - \lambda \nabla_{\mathbf{x}} \|\mathcal{A}\left( \mathbf{\hat{x}_0} \left( \mathbf{x} \right) \right) - \mathbf{y} \|^{2}_{2}$ \Comment{Conditioning on $\mathbf{Y}$ using DPS}
        \State $\mathbf{X_i}[:,:,j] \gets \mathbf{x''}$
    \EndFor
\Else
    \For{$j \text{ in } 1:d_1$} \Comment{Secondary model chosen}
        \State $\mathbf{x} \gets  \mathbf{X_{i+1}}[j,:,:]$
        \State $\mathbf{x'} \gets \texttt{PC sampling} (\mathbf{x}, \mathbf{s_{\phi}}, \sigma_t, t)$
        \State $\mathbf{X_i}[j,:,:] \gets \mathbf{x'}$
    \EndFor
\EndIf
\EndFor \\
\Return $\mathbf{X_0}$
\end{algorithmic}
\end{algorithm}
\section{Experiments}
\subsection{Experimental Settings}
\subsubsection{Dataset} \label{dataset_section}
We acquired a preclinical dataset comprising CBCT scans of ten fresh pig mandibles with preserved soft tissue. The mandibles were not equipped with any dental implants. The scans were performed using both small and large FOVs across four different CBCT scanners: \emph{3D Accuitomo 170} (Morita, Kyoto, Japan), \emph{Axeos} (Dentsply-Sirona, Charlotte, NC, USA), \emph{ProMax 3D Max} (Planmeca, Helsinki, Finland), and \emph{X800} (Morita). Due to the lack of real projection images, scanners geometrical information and for the purposes of the reported experiments we synthesized projection images using TIGRE Toolbox~\cite{biguri2016tigre}. Subsequently, we resampled them to a size of $256 \times 256 \times 256$. The resulting stacks of projection frames were split in two different ways. The stack of projections is a 3D array with indices $i$, $j$, and $k$. Index $i$ enumerates the columns, which are parallel to the rotation axis of the scanner. Index $j$ enumerates the rows, and index $k$ corresponds to the projections at different angles. The $i$–$j$ planes are used for training the primary model. For the secondary model, we chose the $j$–$k$ planes due to their greater resemblance to the $i$–$j$ planes. To make the simulated projections more representative of real-world conditions, we added Poisson and Gaussian noise to our data. The training set comprises nine pig mandibles, while the test set includes one pig mandible. This resulted in a total of $256$ 2D projections $\times $ 9 mandibles $\times $ 4 scanners $\times $ 2 FOVs $=$ \SI{18432} 2D projections for each of the two views for the training set and $256 \times 1 \times 4 \times 2 = $\SI{2048} 2D projections or each of the two views for the test set. Random flips were applied for data augmentation.
\subsubsection{Artefact Simulation}
To validate the method, we inserted synthetic implants into the projection series and simulated the resulting metal artefacts. We created and stored implant masks by placing implants in anatomically plausible locations within the reconstructions, mimicking typical clinical scenarios. We simulated the projections of these implant masks to insert them in the projection series of the mandibles. To introduce artefacts in the reconstruction process, we simulated beam hardening and Poisson noise according to~\cite{zhang2018convolutional}. The details of this simulation are provided in the supplementary material.
\subsubsection{Implementation Details}
We trained two score-based diffusion models with our simulated implant-free projection series on an unconditioned image generation task using the variance-exploding stochastic differential equation, following the approach in~\cite{song2020score}. The main model was trained with images that were projected in a direction along the projection axis. The secondary model was trained with projections in a plane perpendicular to the projection axis, as described in Sect.~\ref{dataset_section}. We chose $T=1000$ noise steps. All experiments were carried out on an NVIDIA A100 (\SI{40}{\giga\byte}) GPU with batch size 8. Both models were trained for 100 epochs, requiring approximately 2.5 days. We modified the implementation provided in~\cite{lee2023improving} to address our inpainting task. Through grid search over the range $(0.001, 0.01, 0.1, 1.0, 10)$, we found that the hyperparameter $\lambda = 1.0$  produced the best results in the sampling process for our dataset. The constant $K$, which controls the frequency ratio between the primary and secondary models, was set to $K=2$, meaning the models alternate at every second denoising step.
\section{Results}
\subsubsection{Comparing Methods} We compared our method TPDM to diffusion posterior sampling in 2D (DPS), where the secondary model was omitted. The same model used as the primary model in the perpendicular diffusion framework was employed for this comparison. Furthermore, we compared our results with a linear interpolation (LI) method using the LinearNDInterpolator from the SciPy Python package~\cite{virtanen2020scipy} that triangulates the input data and performs a linear barycentric interpolation on each triangle. The LI was applied to each 2D projection of the stack individually. Before image reconstruction, we copied the inpainted parts into our original projection series to ensure maximal similarity to the original reconstructions.
\subsubsection{Evaluation} We sampled projection series with two different implant conditions (2 or 3 implants) placed at various locations for all four CBCT scanner units using both large and small-sized FOVs. After sampling all projection series from the test set, we reconstructed the images with the FDK-algorithm~\cite{Feldkamp:84} using the TIGRE Toolbox~\cite{biguri2016tigre}. The images were reconstructed to maintain the original size and resolution of the original reconstructions.

\begin{figure}[h!]
    \centering
        \begin{tikzpicture}[scale=1.18, transform shape]
            \node[] at (1, 4.3) {\scriptsize Artefact};
            \node[] at (3, 4.3) {\scriptsize Reference};
            \node[] at (5, 4.3) {\scriptsize TPDM};
            \node[] at (7, 4.3) {\scriptsize DPS};
            \node[] at (9, 4.3) {\scriptsize LI};
            
            \node[] at (0, 2)   [anchor=south west]  {\includegraphics[height=2cm]{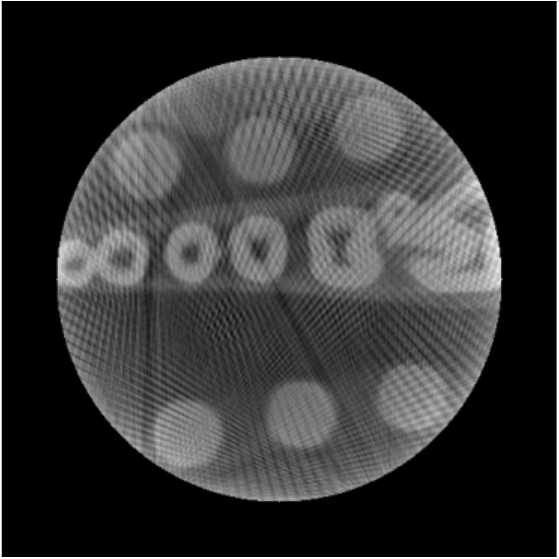}};
            \node[] at (2, 2)   [anchor=south west]  {\includegraphics[height=2cm]{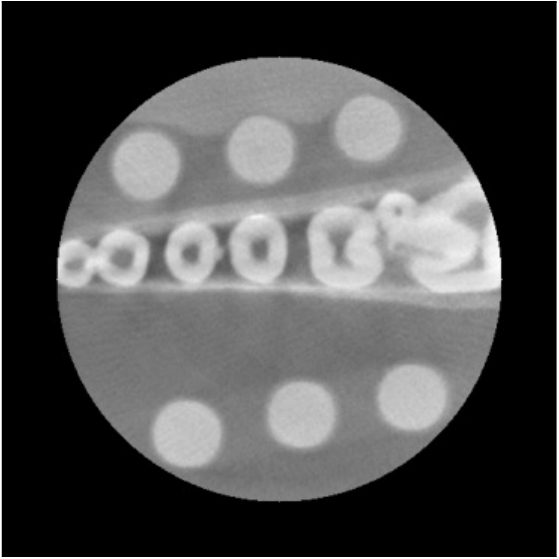}};
            \node[] at (4, 2)   [anchor=south west]  {\includegraphics[height=2cm]{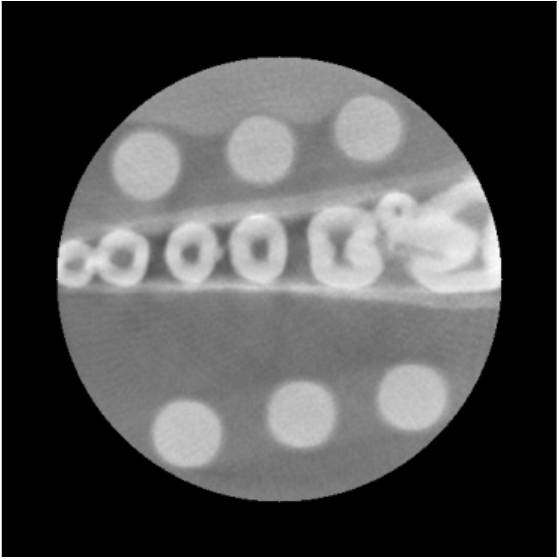}};
            \node[] at (6, 2)   [anchor=south west]  {\includegraphics[height=2cm]{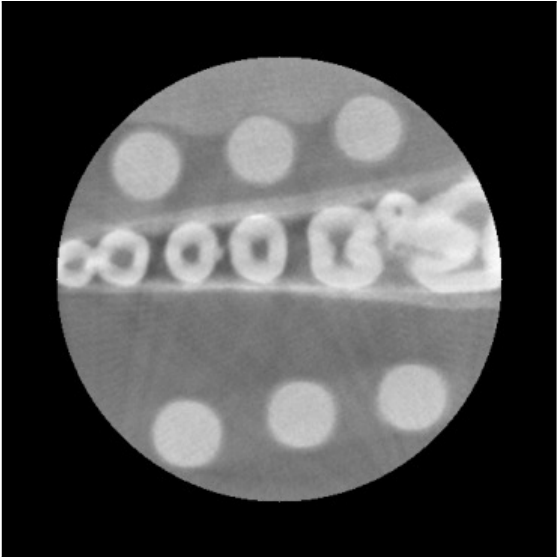}};
            \node[] at (8, 2)   [anchor=south west]  {\includegraphics[height=2cm]{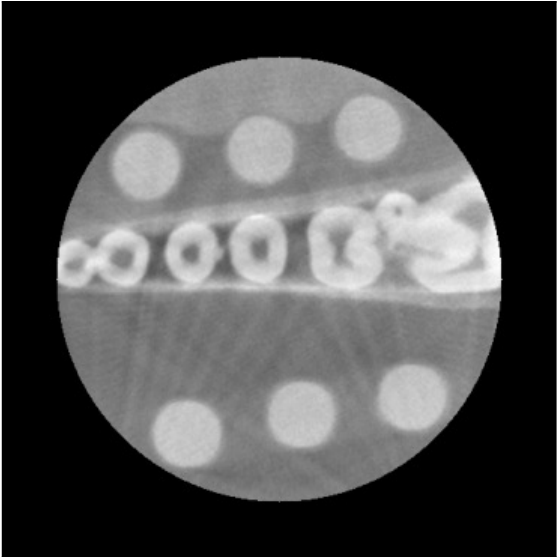}};

            \node[] at (3.4, 0)   [anchor=south west]  {\includegraphics[height=1.9cm]{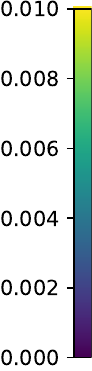}};
            \node[] at (4, 0)   [anchor=south west]  {\includegraphics[height=2cm]{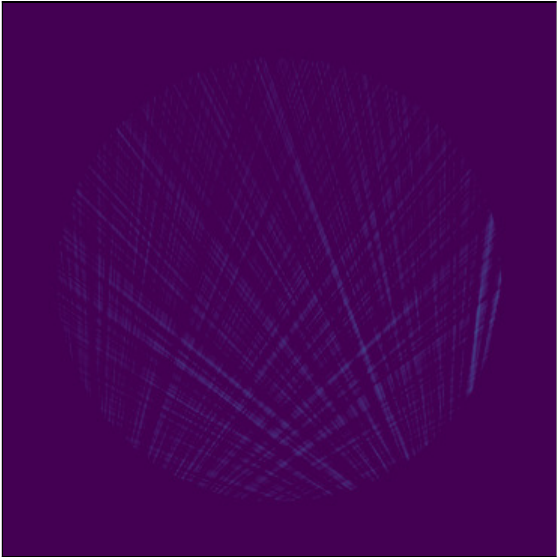}};
            \node[] at (6, 0)   [anchor=south west]  {\includegraphics[height=2cm]{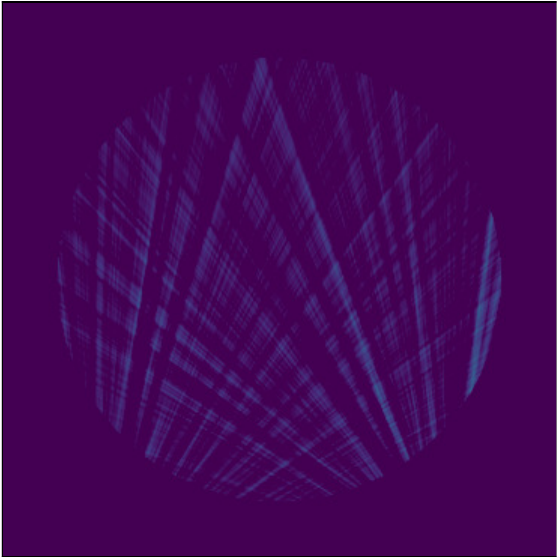}};
            \node[] at (8, 0)   [anchor=south west]  {\includegraphics[height=2cm]{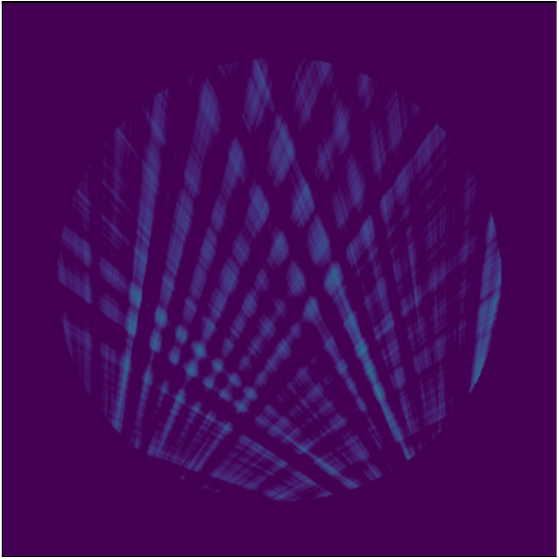}};
        \end{tikzpicture}
    \caption{Top: CBCT reconstructions of an exemplary image. Bottom: Difference maps with respect to the reference image. The images used for the difference maps are in the range [-0.04, 0.07].}
    \label{example_images}
\end{figure}
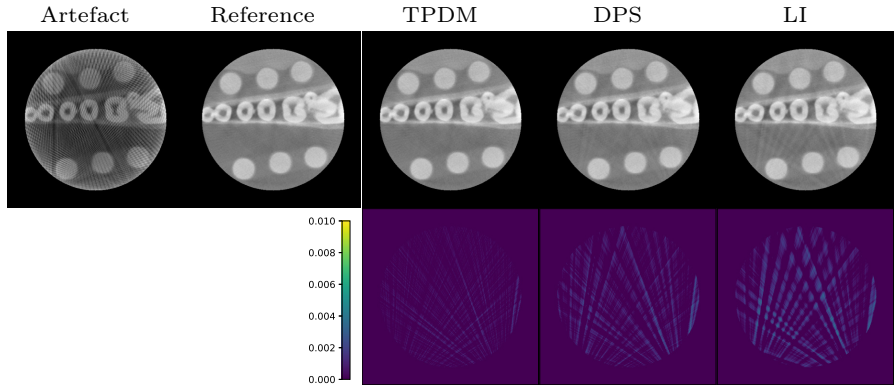

For the evaluation of image quality, we considered root mean square error (RMSE), structural similarity index measure (SSIM), and peak signal-to-noise ratio (PSNR). These metrics were computed within the inpainted regions of the projection series for the artefact-corrected images from the three methods (TPDM, DPS, and LI), using the artefact-free reference image as a baseline. Additionally, to assess artefact reduction in the reconstructed images, we computed these metrics across the reconstructed volumes. The average results are summarised in Table~\ref{tab:evaluation}. We show an exemplary image in Fig.~\ref{example_images}. The image depicts a small FOV, where the implants fall into the exomass and are therefore not visible within the FOV.
\begin{table}[h!]
\centering
\caption{We report the average scores for our test set. The projection row shows the scores calculated only within the inpainted masks in the projections. The results in the reconstruction row were calculated over the whole reconstruction.}
\begin{tabular}{cc|ccccc} \toprule
    &  & SSIM ($\uparrow$) & PSNR ($\uparrow$) & RMSE ($\downarrow$)  \\ \midrule
\multirow{3}{*}{Projections} &TPDM& \bftab 0.9290 $\pm$ 0.0085 & \bftab 35.8660 $\pm$ 1.1841 & \bftab 0.0037 $\pm$ 0.0010 \\
   & DPS& 0.9110 $\pm$ 0.0152 & 33.0039 $\pm$ 1.7146 &  0.0044 $\pm$ 0.0010   \\
   & LI & 0.8834 $\pm$ 0.0168 & 30.3120 $\pm$ 1.1217 & 0.0055 $\pm$ 0.0017  \\\midrule
\multirow{4}{*}{Reconstruction} &TPDM & \bftab 0.9850 $\pm$ 0.0078 & \bftab 50.3875 $\pm$ 3.8834 & \bftab 0.0008 $\pm$ 0.0003 \\
& DPS & 0.9848 $\pm$ 0.0100 & 49.9010 $\pm$ 4.0727  & \bftab 0.0008 $\pm$ 0.0003 \\
   &LI& 0.9810 $\pm$ 0.0084 & 47.1244 $\pm$ 3.3138 & 0.0010 $\pm$ 0.0004 \\ 
   &Artefacts & 0.6320 $\pm$ 0.1064 & 26.0589 $\pm$ 2.4358 & 0.0109 $\pm$ 0.0034 \\\bottomrule
\end{tabular}
\label{tab:evaluation}
\end{table}

With a batch size of 3 in the sampling, TPDM required $\sim\SI{8.5}{h}$ to sample a projection series, utilising around \SI{20}{\giga\byte} of VRAM. The DPS method required for the same task on average $\sim \SI{14}{h}$ while using the same amount of memory. 
\section{Discussion and Conclusion}
In this study, we applied a novel inpainting approach by combining two perpendicular score-based diffusion models to solve the inverse problem of dental implant inpainting in synthetic CBCT projection data in 3D.  Unlike existing implant inpainting techniques, which process 2D projections independently, our approach is the first to extend diffusion model-based dental implant inpainting to the 3D projection stacks to make use of inter-slice correlations.

Our approach has demonstrated strong performance on the diverse test set despite being trained on a relatively small but diverse dataset, which included projection series from nine pig mandibles acquired from four different CBCT units and covered both large- and small-sized FOVs. The method has demonstrated effectiveness in inpainting implants across both large and small FOVs, handling implants located both within and outside the exomass across all four CBCT scanner units, indicating that it holds significant potential for enhancing image quality and diagnostic accuracy in medical imaging.

Our method surpasses both the 2D DPS approach and LI across all metrics. This is particularly evident when computing scores within the inpainted implant masks in the projection series. 
Score differences between TPDM and DPS in the reconstructions are less noticeable, as the FDK algorithm tends to average out errors in the projection sequence.
Both the TPDM and DPS approaches achieve similarly low RMSE values. Nonetheless, the TPDM approach improves the other scores in the reconstructions as well and offers the advantage of a significantly increased sampling speed compared to the DPS method.

The difference maps of the representative image shown in Fig.~\ref{example_images} indicate that the TPDM method yields the smallest deviation from the reference image, supporting the quantitative results.
The results support our hypothesis that inpainting benefits from utilising the entire 3D projection series rather than relying on a single projection.

Furthermore, our approach has the advantage that the training is performed unconditionally, i.e., making it independent of implant mask features.

In this work, scatter modeling was not incorporated. In real CBCT projection data, scatter-corrupted projections can degrade image quality. Previous studies have addressed this challenge using both hardware-based and algorithmic scatter reduction techniques~\cite{cobos20233d,trapp2022empirical,xu2025accelerated}. Our approach could potentially be integrated with these existing methods to enhance image quality.

We were unable to validate our proof of concept on real clinical data due to the lack of publicly available large-scale datasets containing real dental CBCT projections with known scanner geometry. 
Furthermore, full 3D models typically require
significantly more training data than our proposed method. This is because the networks in our proposed method can be trained on the 2D projections which makes them less prone to overfitting. We therefore could not compare our method with full 3D models. 
Future work could involve incorporating a large dataset containing real projections with known scanner geometry to enable 3D reconstruction and facilitate a direct comparison with 3D models.

As sampling from diffusion models is computationally intensive and time-consuming, future work could optimise the process by employing more efficient approaches to improve clinical applicability.
\begin{credits}
\subsubsection{\discintname}
The authors have no competing interests to declare that are
relevant to the content of this article.
\end{credits}
\bibliographystyle{splncs04}
\bibliography{Paper-0043}

@inproceedings{chung2022diffusion,
title={Diffusion Posterior Sampling for General Noisy Inverse Problems},
author={Hyungjin Chung and Jeongsol Kim and Michael Thompson Mccann and Marc Louis Klasky and Jong Chul Ye},
booktitle={The Eleventh International Conference on Learning Representations },
year={2023}
}

@article{biguri2016tigre,
  title={TIGRE: a MATLAB-GPU toolbox for CBCT image reconstruction},
  author={Biguri, Ander and Dosanjh, Manjit and Hancock, Steven and Soleimani, Manuchehr},
  journal={Biomedical Physics \& Engineering Express},
  volume={2},
  number={5},
  pages={055010},
  year={2016},
  publisher={IOP Publishing}
}

@inproceedings{song2020score,
  title={Score-Based Generative Modeling through Stochastic Differential Equations},
  author={Song, Yang and Sohl-Dickstein, Jascha and Kingma, Diederik P and Kumar, Abhishek and Ermon, Stefano and Poole, Ben},
  booktitle={International Conference on Learning Representations},
  year={2021}
}

@article{zhang2018convolutional,
  title={Convolutional neural network based metal artifact reduction in X-ray computed tomography},
  author={Zhang, Yanbo and Yu, Hengyong},
  journal={IEEE transactions on medical imaging},
  volume={37},
  number={6},
  pages={1370--1381},
  year={2018},
  publisher={IEEE}
}

@inproceedings{lee2023improving,
  title={Improving 3D imaging with pre-trained perpendicular 2D diffusion models},
  author={Lee, Suhyeon and Chung, Hyungjin and Park, Minyoung and Park, Jonghyuk and Ryu, Wi-Sun and Ye, Jong Chul},
  booktitle={Proceedings of the IEEE/CVF International Conference on Computer Vision},
  pages={10710--10720},
  year={2023}
}

@article{candemil2019metal,
  title={Are metal artefact reduction algorithms effective to correct cone beam CT artefacts arising from the exomass?},
  author={Candemil, Amanda Pelegrin and Salmon, Benjamin and Freitas, Deborah Queiroz and Ambrosano, Glaucia Maria Bovi and Haiter-Neto, Francisco and Oliveira, Matheus Lima},
  journal={Dentomaxillofacial Radiology},
  volume={48},
  number={3},
  pages={20180290},
  year={2019},
  publisher={Oxford University Press}
}

@article{tang2023improved,
  title={An improved dual-domain network for metal artifact reduction in CT images using aggregated contextual transformations},
  author={Tang, Hui and Jiang, Sudong and Lin, Yubing and Li, Yu and Bao, Xudong},
  journal={Physics in Medicine \& Biology},
  volume={68},
  number={17},
  pages={175021},
  year={2023},
  publisher={IOP Publishing}
}

@inproceedings{lin2019dudonet,
  title={Dudonet: Dual domain network for ct metal artifact reduction},
  author={Lin, Wei-An and Liao, Haofu and Peng, Cheng and Sun, Xiaohang and Zhang, Jingdan and Luo, Jiebo and Chellappa, Rama and Zhou, Shaohua Kevin},
  booktitle={Proceedings of the IEEE/CVF Conference on Computer Vision and Pattern Recognition},
  pages={10512--10521},
  year={2019}
}

@article{karageorgos2024denoising,
  title={A denoising diffusion probabilistic model for metal artifact reduction in CT},
  author={Karageorgos, Grigorios M and Zhang, Jiayong and Peters, Nils and Xia, Wenjun and Niu, Chuang and Paganetti, Harald and Wang, Ge and De Man, Bruno},
  journal={IEEE Transactions on Medical Imaging},
  year={2024},
  publisher={IEEE}
}

@article{wang2024metal,
  title={Metal artifacts reducing method based on diffusion model using intraoral optical scanning data for dental cone-beam CT},
  author={Wang, Yuyang and Liu, Xiaomo and Li, Liang},
  journal={IEEE Transactions on Medical Imaging},
  year={2024},
  publisher={IEEE}
}

@article{oliveira2025development,
  title={Development and evaluation of a deep learning model to reduce exomass-related metal artefacts in cone-beam CT: an ex vivo study using porcine mandibles},
  author={Oliveira, Matheus L and Schaub, Susanne and Dagassan-Berndt, Dorothea and Bieder, Florentin and Cattin, Philippe C and Bornstein, Michael M},
  journal={Dentomaxillofacial Radiology},
  volume={54},
  number={2},
  pages={109--117},
  year={2025},
  publisher={Oxford University Press}
}

@inproceedings{friedrich2024wdm,
  title={Wdm: 3d wavelet diffusion models for high-resolution medical image synthesis},
  author={Friedrich, Paul and Wolleb, Julia and Bieder, Florentin and Durrer, Alicia and Cattin, Philippe C},
  booktitle={MICCAI Workshop on Deep Generative Models},
  pages={11--21},
  year={2024},
  organization={Springer}
}

@article{wang20243d,
  title={3D MedDiffusion: A 3D Medical Diffusion Model for Controllable and High-quality Medical Image Generation},
  author={Wang, Haoshen and Liu, Zhentao and Sun, Kaicong and Wang, Xiaodong and Shen, Dinggang and Cui, Zhiming},
  journal={IEEE transactions on medical imaging},
  year={2025}
}

@article{gottschalk2023dl,
  title={DL-based inpainting for metal artifact reduction for cone beam CT using metal path length information},
  author={Gottschalk, Tristan M and Maier, Andreas and Kordon, Florian and Kreher, Bj{\"o}rn W},
  journal={Medical Physics},
  volume={50},
  number={1},
  pages={128--141},
  year={2023},
  publisher={Wiley Online Library}
}

@article{virtanen2020scipy,
  title={SciPy 1.0: fundamental algorithms for scientific computing in Python},
  author={Virtanen, Pauli and Gommers, Ralf and Oliphant, Travis E and Haberland, Matt and Reddy, Tyler and Cournapeau, David and Burovski, Evgeni and Peterson, Pearu and Weckesser, Warren and Bright, Jonathan and others},
  journal={Nature methods},
  volume={17},
  number={3},
  pages={261--272},
  year={2020},
  publisher={Nature Publishing Group US New York}
}

@article{Feldkamp:84,
author = {L. A. Feldkamp and L. C. Davis and J. W. Kress},
journal = {J. Opt. Soc. Am. A},
number = {6},
pages = {612--619},
publisher = {Optica Publishing Group},
title = {Practical cone-beam algorithm},
volume = {1},
year = {1984}
}

@article{xu2025accelerated,
  title={Accelerated Monte Carlo-driven statistical reconstruction for CBCT scatter correction},
  author={Xu, Wenfeng and An, Guangyan and Yu, Jie and Zhang, Yinghui and Zhao, Xing and Zhang, Huitao and Zhu, Yining},
  journal={Optics Express},
  volume={33},
  number={8},
  pages={18264--18290},
  year={2025},
  publisher={Optica Publishing Group}
}

@article{cobos20233d,
  title={3D-printed large-area focused grid for scatter reduction in cone-beam CT},
  author={Cobos, Santiago Fabian and Norley, Christopher James and Nikolov, Hristo Nikolaev and Holdsworth, David Wayne},
  journal={Medical Physics},
  volume={50},
  number={1},
  pages={240--258},
  year={2023},
  publisher={Wiley Online Library}
}

@article{trapp2022empirical,
  title={Empirical scatter correction: CBCT scatter artifact reduction without prior information},
  author={Trapp, Philip and Maier, Joscha and Susenburger, Markus and Sawall, Stefan and Kachelrie{\ss}, Marc},
  journal={Medical Physics},
  volume={49},
  number={7},
  pages={4566--4584},
  year={2022},
  publisher={Wiley Online Library}
}

@inproceedings{mei2023metal,
  title={Metal inpainting in CBCT projections using score-based generative model},
  author={Mei, Siyuan and Fan, Fuxin and Maier, Andreas},
  booktitle={2023 IEEE 20th International Symposium on Biomedical Imaging (ISBI)},
  pages={1--5},
  year={2023},
  organization={IEEE}
}

\end{document}


\title{Supplementary Material: 3D CBCT Artefact Removal Using Perpendicular Score-Based Diffusion Models}

\author{}
\institute{}
\maketitle              
\section*{Artefact Simulation}
First, we converted the images to Hounsfield units by applying a linear transformation derived from two reference points in the CBCT data, corresponding to air and bone. After conversion, the images were separated into water and bone components, where the water images primarily represent soft tissues and other low-density structures, while the bone images highlight higher-density regions corresponding to osseous structures. We computed the projection series for water and bone images denoted by $\mathbf{p^{w}}$ and $\mathbf{p^{b}}$. We assigned the implant mask a linear attenuation coefficient of a specified metal. The polychromatic projection where beam hardening is simulated is given by~\cite{zhang2018convolutional}:
\begin{equation}
\mathbf{p^{BH}} = \int I \left( E \right) \exp \left( - \frac{m^{w}(E) \mathbf{p^{w}}}{m^{w}(E_0)} - \frac{m^{b}(E) \mathbf{p^{b}}}{m^{b}(E_0)} - \frac{m^{im}(E) \mathbf{p^{im}}}{m^{im}(E_0)} \right) dE
\end{equation}
where $I(E)$ is the intensity as function of the energy, $E_0$ is the monochromatic energy and $m(E)$ is the energy-dependent mass attenuation coefficient. We simulated Poisson noise with
$$ \mathbf{p^{BH,P}} \sim \text{Poisson}\left( \mathbf{p^{BH}}+r \right)$$
where $r$ is the mean number of background events and readout noise variance~\cite{zhang2018convolutional}. The polychromatic projection is then given by:
\begin{equation}
\mathbf{p^{poly}} = - \ln \frac{\mathbf{p^{BH,P}}}{\int I(E) dE}
\end{equation}

\bibliographystyle{splncs04}
\bibliography{Paper-0043}